\documentclass[10pt,twocolumn,letterpaper]{article}

\usepackage{wacv}      

\definecolor{wacvblue}{rgb}{0.21,0.49,0.74}
\usepackage[accsupp]{axessibility}
\usepackage{multirow}
\usepackage{graphicx}
\usepackage[normalem]{ulem}
\useunder{\uline}{\ul}{}
\usepackage{amssymb}
\usepackage{colortbl}
\usepackage[normalem]{ulem}
\usepackage{caption}
\usepackage{stfloats}
\usepackage{subcaption}
\usepackage{float}
\usepackage{pifont}
\usepackage{multirow}
\usepackage{graphicx}
\usepackage{array}
\usepackage{tikz}
\usetikzlibrary{spy}
\usepackage{url}
\usepackage{colortbl}
\usepackage{threeparttable}
\usepackage{placeins}
\usepackage{graphicx,calc}
\usepackage{amsmath}
\usepackage[numbers]{natbib}  
\usepackage[colorlinks=true,linkcolor=blue,citecolor=blue,urlcolor=blue,pagebackref]{hyperref}

\usepackage{booktabs,tabularx,siunitx,makecell}
\def\wacvPaperID{1968} 
\def\confName{WACV}
\def\confYear{2026}

\title{RPT-SR: Regional Prior attention Transformer for infrared image Super-Resolution}

\author{Youngwan Jin$^{1,2}$ \quad Incheol Park$^{1,2}$ \quad Yagiz Nalcakan$^{1}$ \quad Hyeongjin Ju$^{1,2}$ \\ \quad Sanghyeop Yeo$^{1,2}$ \quad Shiho Kim$^{1}$ \\ \\ \quad $^{1}$Yonsei University \\ $^{2}$BK21 Graduate Program in Intelligent Semiconductor Technology  }
\affiliation{organization={\quad Yonsei University}}
\affiliation{organization={\quad BK21 Graduate Program in Intelligent Semiconductor Technology}}
           
\begin{document}
\maketitle
\begin{abstract}
General-purpose super-resolution models, particularly Vision Transformers, have achieved remarkable success but exhibit fundamental inefficiencies in common infrared imaging scenarios like surveillance and autonomous driving, which operate from fixed or nearly-static viewpoints. These models fail to exploit the strong, persistent spatial priors inherent in such scenes, leading to redundant learning and suboptimal performance. To address this, we propose the Regional Prior attention Transformer for infrared image Super-Resolution (RPT-SR), a novel architecture that explicitly encodes scene layout information into the attention mechanism. Our core contribution is a dual-token framework that fuses (1) learnable, regional prior tokens, which act as a persistent memory for the scene's global structure, with (2) local tokens that capture the frame-specific content of the current input. By utilizing these tokens into an attention, our model allows the priors to dynamically modulate the local reconstruction process. Extensive experiments validate our approach. While most prior works focus on a single infrared band, we demonstrate the broad applicability and versatility of RPT-SR by establishing new state-of-the-art performance across diverse datasets covering both Long-Wave (LWIR) and Short-Wave (SWIR) spectra. Code is available at https://github.com/Yonsei-STL/RPT-SR.git.
\end{abstract}
    
\section{Introduction}
\label{sec:intro}

\begin{figure}[t!]
    \centering
    \includegraphics[width=1\linewidth]{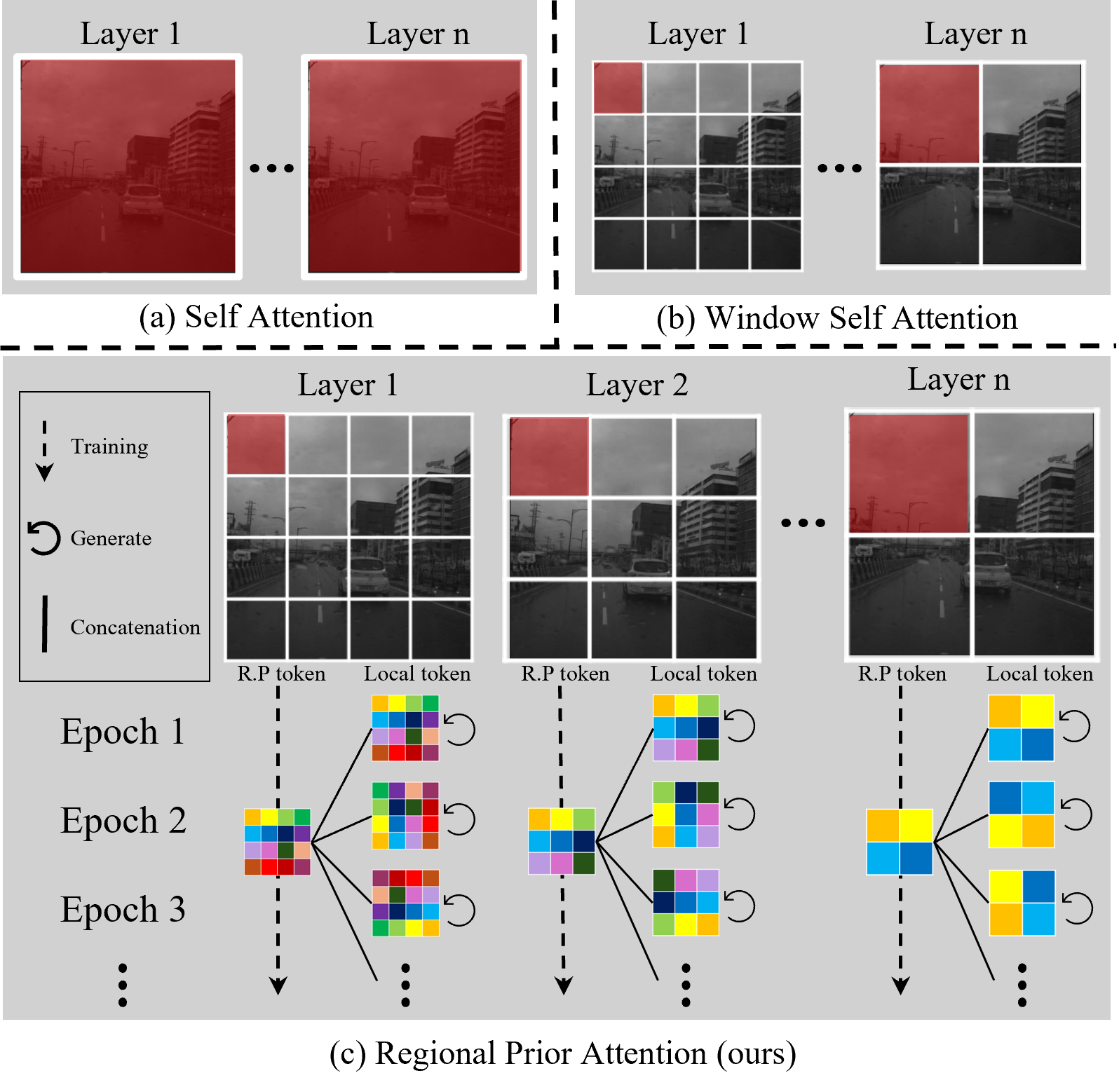}
    \caption{Comparison of attention mechanisms. (a) Standard self-attention computes global relationships at a high computational cost. (b) Window self-attention limits computation to local windows but misses global context. (c) Our proposed Regional Prior Attention (RPA) fuses a persistent, learnable Regional Prior (R.P.) token with a Local token. The R.P. token learns the scene's static layout over epochs, providing a strong structural guide for the reconstruction.}
    \label{fig:summary}
\end{figure}

Infrared (IR) imaging, which operates beyond the visible spectrum, is a key modality that expands the perceptual capabilities of computer vision systems. In particular, the long-wave infrared (LWIR)~\cite{kaist, kaist_ms2} and short-wave infrared (SWIR)~\cite{rasmd, swir_surveillance} bands play indispensable roles across diverse applications because of their distinct physical properties~\cite{driggers2013good}. LWIR (approximately 7 to 14~$\mu$m) senses thermal radiation emitted by objects, enabling illumination-invariant, around-the-clock observation. By contrast, SWIR (approximately 1 to 2.5~$\mu$m) measures reflected light, similar to the visible spectrum, yet exhibits higher transmittance through atmospheric scatterers such as fog, smoke, and haze, thereby delivering sharp images even under adverse weather conditions. These characteristics overcome the limitations of conventional visible (RGB) cameras, whose performance degrades severely in challenging conditions such as fog, rain, glare, and nighttime, and establish IR imaging as a cornerstone of all-weather, robust perception systems~\cite{hyper_driving}.

Infrared (IR) imaging provides unique physical information that visible-light sensors cannot capture, such as material-level distinctions in SWIR~\cite{swir_material} or thermal patterns in LWIR. However, IR sensors face a fundamental limitation of low resolution. This is due to physical and economic constraints—unlike with visible-light sensors—where manufacturing costs rise exponentially with resolution~\cite{s24144686, s23218717, su141811161}. Super-resolution (SR) is therefore not merely for improving image quality, but is the most practical and cost-effective solution to replace expensive high-resolution hardware. Consequently, for advanced vision systems requiring high reliability, such as autonomous driving and remote surveillance, the high-resolution restoration of IR images is of decisive importance~\cite{huang2025infraredimagesuperresolutionsystematic, rs16214033}.

Deep learning-based super-resolution (SR) has achieved remarkable progress in recent years. Early convolutional models such as SRCNN~\cite{srcnn} outperformed classical methods, yet the locality of convolutional kernels limits their ability to model long-range dependencies and global context. To address these limitations, vision transformers have introduced a new paradigm for image restoration. State-of-the-art models such as SwinIR~\cite{swinir} and HAT~\cite{HAT} capture global relationships among image patches via self-attention and achieve results that substantially surpass prior benchmarks. Nevertheless, despite their strong performance, these general-purpose SR models exhibit fundamental inefficiencies in scenarios where IR imaging is frequently deployed, including traffic CCTV, roadside surveillance, and vehicle-mounted ADAS cameras, which operate from fixed or nearly static viewpoints. In such environments, scene structure is highly predictable. For example, in forward-facing driving videos, the road consistently appears at the bottom of the frame, buildings in the upper middle, and the sky at the top. A strong anisotropic spatial prior exists in which the statistical distribution of image content (frequencies, textures, and object classes) changes systematically with pixel location. Existing CNN- and transformer-based SR models do not explicitly encode this layout prior. Instead, during training they must implicitly relearn the same spatial regularities from data, which wastes part of the model’s attention budget on low-information regions and slows convergence. The strength of modern transformers, namely the capacity to model global context dynamically for arbitrary inputs, can become a liability in static-viewpoint settings. Because the model remains in a state of structural amnesia with respect to persistent scene layout, it expends significant capacity rediscovering redundant information in every frame. Consequently, a powerful but statistically naive model allocates resources inefficiently in environments with stable regularities. 

To address this inefficiency, we propose the \textit{Regional Prior attention Transformer for infrared image Super-Resolution (RPT-SR)}. As illustrated in Figure~\ref{fig:summary} (c), our approach introduces a novel attention mechanism called Regional Prior Attention (RPA). The core idea is a dual-token framework that operates on two distinct types of information carriers. First, a learnable, static \textit{Regional Prior (R.P.) token} acts as a persistent memory, learning the scene's invariant spatial layout across the entire dataset over training epochs. Second, a dynamic \textit{Local token} is generated from each input image to capture its unique, frame-specific content. These two tokens are then fused and injected into the attention block, allowing the powerful, scene-constant prior to modulate and guide the reconstruction of specific local details. This design enables the model to escape the "structural amnesia" of prior methods, leading to more efficient learning and higher fidelity. To substantiate the robustness and generalizability of our proposed method, we validate it across two physically distinct infrared spectra. Long-Wave Infrared (LWIR) imaging captures emitted thermal radiation, while Short-Wave Infrared (SWIR) imaging relies on reflected light, leading to fundamentally different image characteristics, noise profiles, and texture distributions. By achieving state-of-the-art results in both domains—a broader validation than is typical in prior work—we demonstrate that our regional prior mechanism is not merely tailored to a specific modality. Instead, it effectively learns the underlying structural regularities of fixed-viewpoint scenes regardless of the imaging physics, confirming the versatility of our approach. The contributions of this paper are as follows: \begin{itemize}
    \item We introduce Regional Prior Attention (RPA), a novel attention mechanism implemented via an effective dual-token architecture. This framework fuses persistent, static prior tokens with frame-specific, dynamic local tokens to explicitly encode the spatial priors of fixed-viewpoint scenes.
    \item We demonstrate the broad applicability and versatility of our method by achieving new state-of-the-art results on diverse benchmarks covering both Long-Wave (LWIR) and Short-Wave (SWIR) infrared spectra.
\end{itemize}

\section{Related Work}
\label{sec:formatting}

\subsection{GAN-Based Image Restoration}
Early advancements in perceptual super-resolution (SR) were largely driven by generative adversarial networks (GANs). ESRGAN\cite{esrgan} coupled a perceptual loss with a relativistic discriminator to enhance visual fidelity, setting a strong baseline for GAN-based methods. However, its degradation model was too simple for real-world scenarios. To address this, BSRGAN\cite{bsrgan} introduced a more practical degradation model featuring a shuffled degradation strategy that randomly mixes complex blur, resampling, and noise types. This allowed it to achieve superior perceptual quality on real photographs, as measured by lower NIQE scores. Building on this, Real-ESRGAN\cite{realesrgan} further enlarged the corruption space and redesigned the generator and discriminator, demonstrating that a well-designed model could be trained for real-world blind SR using purely synthetic data. Although these works showcase the power of GANs, they often depend on hand-crafted degradation pipelines and can suffer from adversarially unstable training, motivating the exploration of alternative generative frameworks.

\subsection{Transformer-based Image Restoration}
The success of self-attention has made Vision Transformers a dominant architecture for image restoration. SwinIR~\cite{swinir} adapts the Swin Transformer to restoration by computing self-attention within shifted local windows, achieving strong performance while avoiding the quadratic cost of global attention. A key challenge in this line of work is how to expand the effective receptive field. To this end, CAT~\cite{cat} enlarges the attention area using parallel horizontal and vertical rectangular windows (Rwin-SA), and ART~\cite{art} alternates between dense local attention and sparse attention blocks that sample tokens from distant locations.

Another direction integrates global information or refines the attention mechanism itself. ATD~\cite{atd} introduces a learnable token dictionary and injects external priors via cross-attention, whereas RGT~\cite{rgt} aggregates features into a compact representative map for efficient global exchange. PFT-SR~\cite{pft-sr} proposes Progressive Focused Attention, which multiplicatively inherits attention maps across layers to emphasize relevant tokens and skip computation on irrelevant ones. Hybrid designs such as HAT~\cite{HAT} and DAT~\cite{dat} further improve representation power by combining channel attention with transformer blocks and by learning data-dependent deformable attention patterns.

Efficiency-oriented models aim to retain quality under limited computation. HiT-SR~\cite{hitsr} replaces shifted-window attention with a spatial–channel correlation mechanism within a hierarchical backbone, SRFormer~\cite{srformer} uses permuted self-attention to aggregate information over larger regions at low cost, and SMFANet~\cite{smfanet} adopts a parallel architecture that explicitly models both local and non-local features to achieve a favorable performance–complexity trade-off.

\subsection{Diffusion-Based Super-Resolution}
Score-based diffusion models have emerged as a stable alternative to adversarial training. SR3~\cite{sr3} demonstrated high-quality SR with a noise-conditioned U-Net but required hundreds of reverse steps, motivating faster sampling. ResShift~\cite{resshift} operates in residual space to reduce sampling to $\sim$15 steps, while SinSR~\cite{sinsr} distills the multi-step reverse process into a single deterministic mapping via a consistency-based loss. For model compression, Bi-DiffSR~\cite{bidiffsr} investigates an ultra-compressed binarized U-Net with dedicated modules (CP-Down/Up and timestep-aware redistribution).

Diffusion backbones have also evolved beyond U-Nets: DiT-SR~\cite{ditsr} proposes an isotropic Transformer that reallocates capacity toward high-resolution stages, and Inf-DiT~\cite{infdit} introduces memory-efficient block attention to enable over-4K generation. For infrared SR, DifIISR~\cite{difiisr} further injects thermal spectrum priors and detector gradients into the diffusion process to improve both visual quality and downstream detection performance.

\subsection{Frequency and Prompt-Guided Approaches}
Several methods improve restoration by incorporating explicit frequency-domain priors or external guidance from text prompts. CoRPLE\cite{corple} enhances structural fidelity in infrared SR by combining a Laplacian-pyramid contourlet transform with guidance from positive and negative text prompts. The contourlet transform helps preserve multi-scale and multi-directional details, while the prompts guide the model toward desired attributes and away from artifacts. Such cues are largely orthogonal to the choice of the generative backbone and can be integrated with Transformer or diffusion-based frameworks to further boost performance.

\section{Method}
\label{sec:method}
\begin{figure*}[ht!]
    \centering
    \includegraphics[width=1\linewidth]{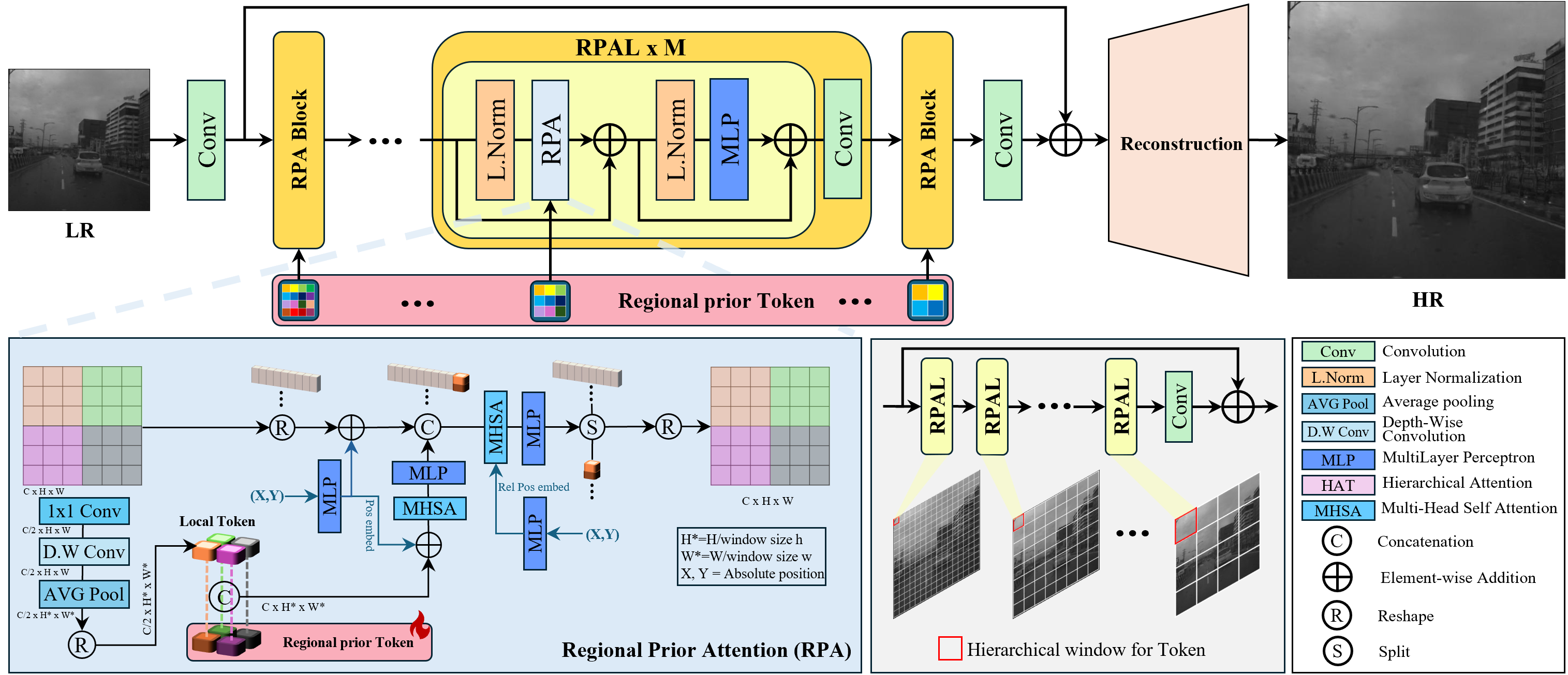}
    \caption{The overall architecture of our proposed RPT-SR. (Top) The model consists of a shallow feature stem, a deep body of RPA Blocks, and a reconstruction head. (Bottom Left) A detailed view of the Regional Prior Attention (RPA) module. A dynamic Local Token, summarized from the input, is fused with a learnable, static Regional Prior Token. These are processed by a Hierarchical Attention (HAT) block to guide the reconstruction. (Bottom Right) The hierarchical windowing strategy, where the attention window size increases in deeper layers.}
    \label{fig:framework}
\end{figure*}

Given a single–frame low‑resolution (LR) infrared image
$\mathit{I}_{\text{LR}}\!\in\!\mathbb{R}^{3 \times H \times W}$,
our goal is to predict its high‑resolution (HR) counterpart $\hat{\mathit{I}}_{\text{HR}}
 \in \mathbb{R}^{3 \times rH \times rW}$. We support an arbitrary up‑sampling factor $r$.
We first motivate our design (Sec. \ref{sec:motivation}), then outline the architecture (Sec. \ref{sec:arch}), and finally formalize the proposed \emph{Regional Prior Attention Layer} (Sec. \ref{sec:rpal}).

\subsection{Motivation}
\label{sec:motivation}

\paragraph{Layout‑conditioned low‑level statistics.}
Fixed‑view infrared imaging (e.g., fixed surveillance, forward‑facing automotive cameras)
exhibits strong \emph{spatial anisotropy} in \emph{low‑level} statistics:
across frames captured from nearly identical geographic structure, poses, the distribution of frequency content, thermal contrast, and texture density vary systematically with pixel location.
Formally, for a low‑level statistic $S$ on image $I$,
\[
\operatorname{Var}_{(u,v)}\,\mathbb{E}_{I}\big[S(I)[u,v]\big] \;\gg\; 0,
\]
indicating non‑uniform, position‑conditioned statistics.
Conventional CNN and Transformer SR backbones do not encode this layout prior explicitly. They must relearn the same spatial regularities from scratch, which slows convergence and dilutes attention over consistently less informative regions.

\paragraph{Regional priors as a compact memory.}
We operationalize these class‑agnostic, layout‑conditioned priors by introducing a learnable, spatially indexed memory called \textit{regional prior token}, which is utilized per macro‑window location. These tokens are shared across images and optimized end‑to‑end using only the pixel reconstruction loss to accelerate their early learning. We apply a learning-rate multiplier to these parameters only. At run time, the regional prior tokens are paired with per‑input local tokens distilled from the current feature map; concatenating the two along channels yields dynamic tokens that inject global, position‑specific regularities into attention. This design accelerates convergence and improves reconstruction fidelity on fixed‑view IR scenes.

\subsection{Network Overview}
\label{sec:arch}

Figure~\ref{fig:framework} shows the overall pipeline, following the standard ``feature extractor $\rightarrow$ deep transformer body $\rightarrow$ reconstruction head'' pattern used in transformer‑based SR. \textbf{(1) Shallow feature stem.}
A $3{\times}3$ convolution maps $\mathit{I}_{\text{LR}}$ to
$\mathit{F}_0\!\in\!\mathbb{R}^{C\times H\times W}$ without changing resolution.
No absolute positional encoding is used here, preserving input‑size agnosticism. \textbf{(2) Deep body of Regional‑Prior Attention (RPA) blocks.}
The core is a cascade of residual \emph{RPA Blocks}. Each block contains several RPA layers. In each RPA layer, window‑based tokens (the usual per‑pixel tokens inside a window) interact with a small set of dynamic tokens formed by concatenating a per‑input local token and a learnable regional prior token (RPT) at the same macro‑window location. Inside every layer, we design a regional prior attention based on FasterViT~\cite{fastervit} to refine the dynamic tokens through a lightweight self‑attention and inject the refined dynamic tokens into the window attention, so that global, position‑specific regularities modulate local interactions. Residual paths and an MLP complete the layer; several layers at the same resolution are wrapped by an outer residual path to form one RPA Block. \textbf{(3) Reconstruction head.} After the deep body, a $3{\times}3$ aggregation convolution is applied. A single pixel‑shuffle unit upsamples by $r$, and a final $3{\times}3$ convolution produces $\hat{\mathit{I}}_{\text{HR}}$.

\subsection{Regional Prior Attention}
\label{sec:rpal}
\noindent\textbf{Notation.}
Let $N_w=\frac{HW}{w^2}$ be the number of windows and $k$ the number of dynamic tokens per window,
so the total number of dynamic tokens is $N_c=k\,N_w$.\\

\noindent Let $\mathit{F}\!\in\!\mathbb{R}^{C\times H\times W}$ be the input feature map to one RPA,
and let $w$ denote the window edge length.

\paragraph{1) Window tokenization.}
Partition $\mathit{F}$ into $N_w=\tfrac{HW}{w^2}$ windows. For each window
$j\!\in\!\{1,\dots,N_w\}$ we form the \emph{window token sequence}
\[
\mathit{X}^{(j)} \in \mathbb{R}^{w^{2}\times C},
\tag{1}
\]
i.e., \emph{$w^2$ tokens per window}.

\paragraph{2) Local token generation (per input).}
A lightweight function $\phi$ first reduces channels with a $1{\times}1$ projection
($C\!\to\!C/2$), followed by a depth‑wise $3{\times}3$ convolution and average pooling,
producing a coarse grid of \emph{local tokens}
\[
\mathit{L}=\phi(\mathit{F})
\in \mathbb{R}^{N_c\times C/2}.
\tag{2}
\]
We denote by $k$ the number of tokens assigned to each window,
so that the total number is $N_c = k\,N_w$.\footnote{Our experiments use $k{=}1$;
general $k$ is supported by the implementation.}

\paragraph{3) Regional prior token (learned parameter).}
Each macro‑window location is equipped with a learnable \emph{regional prior token}
\[
\mathit{R}\in\mathbb{R}^{N_c\times C/2},
\tag{3}
\]
initialized from the first mini‑batch (by copying the corresponding local tokens) and
then optimized by SGD. $\mathit{R}$ is shared across images but \emph{distinct per layer}
(i.e., not shared across layers).

\paragraph{4) Dynamic token (fusion).}
At each location, the local token and regional prior token are concatenated along channels
to form a \emph{dynamic token}
\[
\mathit{D}
=
\bigl[\mathit{L} \;\|\; \mathit{R}\bigr]
\in \mathbb{R}^{N_c\times C}.
\tag{4}
\]
No extra gating scalar is used.

\paragraph{5) attention with dynamic tokens.}
attention proceeds in two stages:

\textbf{(a) Dynamic‑token self‑attention.}
Dynamic tokens exchange information globally:
\begin{align}
\tilde{\mathit{D}}
&=
\operatorname{MSA}\!\bigl(\operatorname{LN}(\mathit{D})\bigr)
+
\mathit{D},
\tag{5}\\
\mathit{D}^{\ast}
&=
\operatorname{MLP}\!\bigl(\operatorname{LN}(\tilde{\mathit{D}})\bigr)
+
\tilde{\mathit{D}}.
\tag{6}
\end{align}

\textbf{(b) Window attention with dynamic tokens.}
The refined dynamic tokens are redistributed so that each window
receives $k$ of them. For window $j$ we prepend the $k$ dynamic tokens and run
attention over the concatenated sequence:
\[
\mathit{Z}^{(j)}=\bigl[\mathit{D}^{\ast,(j)} \;\|\; \mathit{X}^{(j)}\bigr]
\in \mathbb{R}^{(k+w^2)\times C},
\]
\[
\mathit{Y}^{(j)}
=
\operatorname{MSA}\!\bigl(\operatorname{LN}(\mathit{Z}^{(j)})\bigr)
+
\mathit{Z}^{(j)},\]
\[
\qquad
\mathit{O}^{(j)}
=
\operatorname{MLP}\!\bigl(\operatorname{LN}(\mathit{Y}^{(j)})\bigr)
+
\mathit{Y}^{(j)}.
\tag{7}
\]
The outputs $\{\mathit{O}^{(j)}\}_{j=1}^{N_w}$ are reshaped back to
\paragraph{Complexity.}
Per window, the attention cost scales as $\mathcal{O}((k+w^2)^2 C)$; relative to a $k{=}0$ baseline, 
the incremental overhead is $\mathcal{O}(2kw^2 + k^2)$, which stays small for the $k\!\le\!4$ used in practice.
$\mathbb{R}^{C\times H\times W}$.

\subsection{Implementation Details}\label{sec:impl}

For the classical model (RPT-SR), we use 4 RPA blocks, each containing 4 transformer layers with multi-head attention(6 heads). The embedding dimension is 240 channels, and the hierarchical window sizes follow the schedule 
$[8{\times}8,\,16{\times}16,\,16{\times}16,\,32{\times}32]$. We use one learnable dynamic token per attention window ($k{=}1$) and optimize the model with Adam ($(\beta_1{=}0.9,\ \beta_2{=}0.99)$) at an initial learning rate of $5{\times}10^{-4}$ for $100$k iterations under a MultiStep schedule (milestones at $50$k/$80$k/$90$k/$92.5$k, $\gamma{=}0.5$). Regional prior tokens are trained with a $50\times$ learning rate multiplier, while all other parameters use the base rate. For the lightweight variant (RPT-SR-Light), we reduce the configuration to 4 heads and 80 channels, with window sizes $[8{\times}8,\,8{\times}8,\,16{\times}16,\,16{\times}16]$), while keeping all other training settings (loss, optimizer, schedule, and $k{=}1$) identical to the baseline model.

\begin{table*}[ht!]
  \centering
  \setlength{\tabcolsep}{9pt} 
  \vspace{2mm}
  \begin{tabular*}{\textwidth}{
    @{\extracolsep{\fill}}
    l l
    S[table-format=1.4]
    S[table-format=2.4]
    S[table-format=1.4]
    S[table-format=3.4]
    S[table-format=2.3]
  }
    \toprule
    \textbf{Method} & \textbf{Publication} &
    \multicolumn{1}{c}{\textbf{LPIPS}\,$\downarrow$} &
    \multicolumn{1}{c}{\textbf{MUSIQ}\,$\uparrow$} &
    \multicolumn{1}{c}{\textbf{MANIQA}\,$\uparrow$} &
    \multicolumn{1}{c}{\textbf{FLOPs (G)}\,$\downarrow$} &
    \multicolumn{1}{c}{\textbf{Params (M)}\,$\downarrow$} \\
    \midrule
    BSRGAN~\cite{bsrgan}         & CVPR'21        & 0.237155 & 33.61297 & 0.164349 & 297.730 & 16.700 \\
    SwinIR~\cite{swinir}         & ICCVW'21       & 0.206968 & 31.85501 & 0.144053 & 192.080 & 11.900 \\
    HAT~\cite{HAT}            & CVPR'23        & 0.111827 & 39.62691 & 0.244789 & 345.632 & 20.510 \\
    HAT-s~\cite{HAT}          & CVPR'23        & 0.111712 & 40.07024 & 0.243761 & 163.313 &  9.359 \\
    DAT~\cite{dat}            & ICCV'23        & 0.108384 & 40.24581 & 0.247306 & 245.182 & 14.802 \\
    DAT-light~\cite{dat}      & ICCV'23        & 0.158623 & 38.16357 & 0.238152 &   9.106 &  0.600 \\
    DAT-S~\cite{dat}          & ICCV'23        & 0.114782 & 40.08282 & 0.246773 & 186.468 & 11.212 \\
    DAT2~\cite{dat}           & ICCV'23        & 0.142401 & 40.2149  & 0.248835 & 186.471 & 11.212 \\
    RGT~\cite{rgt}            & ICLR'24        & 0.109421 & 41.30493 & 0.248173 & 207.604 & 13.357 \\
    RGT-S~\cite{rgt}          & ICLR'24        & 0.124055 & 39.61056 & 0.243492 & 160.481 & 10.192 \\
    CAT-a2~\cite{cat}         & NeurIPS'22     & 0.107230 & 40.84209 & 0.244462 & 278.923 & 16.604 \\
    CAT-a~\cite{cat}          & NeurIPS'22     & 0.110688 & 40.12915 & 0.243888 & 278.918 & 16.604 \\
    CAT2~\cite{cat}           & NeurIPS'22     & 0.118609 & 39.26446 & 0.239382 & 202.467 & 11.925 \\
    CAT~\cite{cat}            & NeurIPS'22     & 0.133433 & 41.73532 & 0.2473206 & 278.908 & 16.604 \\
    SMFANet~\cite{smfanet}        & ECCV'24        & 0.168162 & 36.66893 & 0.234465 &   2.949 &  0.197 \\
    SMFANet-plus~\cite{smfanet}    & ECCV'24        & 0.165560 & 37.15005 & 0.246253 &   7.447 &  0.495 \\
    SRFormer~\cite{srformer}        & ICCV'23        & 0.121545 & 40.00695 & 0.245516 & 227.404 & 10.429 \\
    SRFormer-light~\cite{srformer} & ICCV'23        & 0.173939 & 35.09660 & 0.233217 &  14.134 &  0.840 \\
    ART~\cite{art}            & ICLR'23        & 0.114752 & 39.06023 & 0.239016 & 283.757 & 16.546 \\
    ART-S~\cite{art}          & ICLR'23        & 0.180067 & 37.66906 & 0.240764 & 203.214 & 11.867 \\
    PFT~\cite{pft-sr}            & CVPR'25        & \underline{\num{0.104529}} & 38.98873 & 0.245096 & 319.166 & 19.057 \\
    PFT-light~\cite{pft-sr}      & CVPR'25        & 0.157731 & 37.58969 & 0.242331 &  11.337 &  0.697 \\
    ATD~\cite{atd}            & CVPR'24        & 0.135661 & 41.67953 & \underline{\num{0.249601}}& 310.235 & 19.884 \\
    ATD-light~\cite{atd}      & CVPR'24        & 0.162549 & 37.47317 & 0.248585 &  11.669 &  0.661 \\
    HIT-SR~\cite{hitsr}         & ICCV'24 oral   & 0.151644 & 38.35157 & 0.237546 &  13.066 &  0.792 \\
    CoRPLE~\cite{corple}         & ECCV'24        & 0.232045 & 31.07335 & 0.202224 &  37.592 &  0.591 \\
    \midrule
    \textbf{RPT-SR-light (Ours)} & & 0.137673 & \underline{\num{41.73719}} & 0.247214 & 27.554 & 3.950 \\
    \textbf{RPT-SR (Ours)} & 
    & \textbf{\num{0.103793}}
    & \textbf{\num{41.8049}}
    & \textbf{\num{0.26213}}
    & \num{237.783}
    & 25.830 \\
    \bottomrule
  \end{tabular*}
  \caption{Quantitative comparison for $\times4$ super-resolution on the M3FD dataset. The best results are marked in \textbf{bold} and the second-best are \underline{underlined}.}
  \label{tab:m3fd}
\end{table*}

\section{Experiments}

\subsection{Evaluation Metrics}
\label{sec:metrics}
\begin{figure*}[ht!]
    \centering
    \includegraphics[width=1\linewidth]{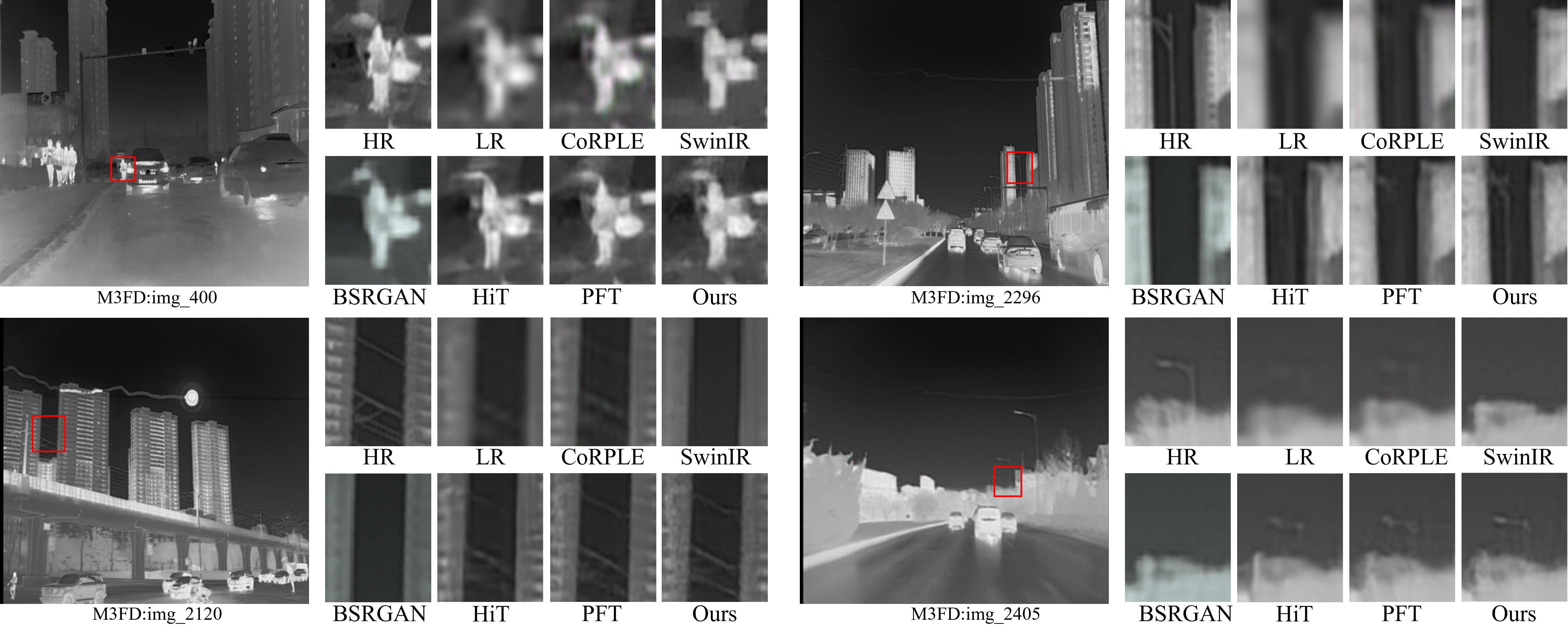}
    \caption{Qualitative comparison for $\times$4 super-resolution on the M3FD dataset. Our method (RPT-SR) reconstructs sharper details and more plausible textures compared to existing state-of-the-art methods, particularly in restoring fine structures like human figures, building facades, and distant objects.}
    \label{fig:vis}
\end{figure*}
Traditional metrics such as Peak Signal-to-Noise Ratio (PSNR) and Structural Similarity Index (SSIM) are widely used but are not reliable indicators of perceptual quality. Prior work in super-resolution has shown that these pixel-level metrics often favor blurry regression outputs that achieve higher scores but diverge from human preference~\cite{photorealistic, pixelrecursive, pulse}. Similarly, Palette~\cite{palette} emphasizes that such metrics fail to reflect the quality of generated samples in hallucination-heavy tasks. To obtain a more perceptually aligned evaluation, we adopt a suite of complementary metrics. For reference-based fidelity, we use the Learned Perceptual Image Patch Similarity (LPIPS)~\cite{lpips}, which measures distances in deep feature space and shows strong correlation with human judgment. For no-reference assessment, we employ MUSIQ~\cite{musiq}, a Transformer-based model that aggregates multi-scale features to predict perceptual quality. In addition, to capture distortions and artifacts common in tasks such as super-resolution, we incorporate MANIQA~\cite{maniqa}, a state-of-the-art no-reference IQA method.

\begin{table}[h!]
  \centering
  \footnotesize
  \vspace{2mm}
  \begin{tabularx}{\linewidth}{%
    >{\raggedright\arraybackslash}X
    S[table-format=1.6]
    S[table-format=1.6]
  }
    \toprule
    \textbf{Method} & \multicolumn{2}{c}{\textbf{LPIPS}\,$\downarrow$} \\
    \cmidrule(lr){2-3}
    & {\textbf{RASMD}} & {\textbf{TNO}} \\
    \midrule
    BSRGAN~\cite{bsrgan}        & \num{0.188441} & \num{0.409504} \\
    SwinIR~\cite{swinir}        & \num{0.241546} & \num{0.479653} \\
    HAT~\cite{HAT}              & \num{0.156022} & \textbf{\num{0.247531}} \\
    HAT-S~\cite{HAT}            & \num{0.160292} & \num{0.254256} \\
    DAT~\cite{dat}              & \num{0.162312} & \num{0.253371} \\
    DAT\_light~\cite{dat}       & \num{0.165045} & \num{0.353354} \\
    DAT-S~\cite{dat}            & \num{0.160324} & \num{0.254787} \\
    RGT~\cite{rgt}              & \num{0.163848} & \num{0.280855} \\
    RGT-S~\cite{rgt}            & \num{0.158238} & \num{0.285347} \\
    CAT~\cite{cat}              & \num{0.155913} & \num{0.327957} \\
    CAT-2~\cite{cat}            & \num{0.160428} & \num{0.297327} \\
    CAT-A~\cite{cat}            & \underline{\num{0.154804}} & \num{0.327957} \\
    CAT-A2~\cite{cat}           & \num{0.15774 } & \num{0.326901} \\
    SMFANet~\cite{smfanet}      & \num{0.175476} & \num{0.360945} \\
    SMFANet-plus~\cite{smfanet} & \num{0.171053} & \num{0.357373} \\
    ART~\cite{art}              & \num{0.159169} & \num{0.253610} \\
    ART-S~\cite{art}            & \num{0.164977} & \num{0.337422} \\
    PFT~\cite{pft-sr}              & \num{0.160775} & \num{0.322897} \\
    PFT-light~\cite{pft-sr}        & \num{0.164730} & \num{0.343978} \\
    ATD~\cite{atd}              & \num{0.158876} & \num{0.316261} \\
    ATD-light~\cite{atd}        & \num{0.166650} & \num{0.343077} \\
    HIT-SR~\cite{hitsr}         & \num{0.167474} & \num{0.344218} \\
    CoRPLE~\cite{corple}        & \num{0.308662} & \num{0.367111} \\
    \midrule
    \textbf{RPT-SR (Ours)}  & \textbf{\num{0.15354}} & \underline{\num{0.250140}} \\
    \bottomrule
  \end{tabularx}
  \caption{Comparison of LPIPS scores on RASMD and TNO datasets ($\times4$).}
  \label{tab:ablation-rasmd}
\end{table}

\begin{table}[h!]
  \centering
  \footnotesize
  \vspace{2mm}
  \begin{tabularx}{\linewidth}{%
    >{\raggedright\arraybackslash}X
    S[table-format=1.6]
    S[table-format=1.6]
  }
    \toprule
    \textbf{Method} & \multicolumn{2}{c}{\textbf{LPIPS}\,$\downarrow$} \\
    \cmidrule(lr){2-3}
    & {\textbf{RASMD}} & {\textbf{M3FD}} \\
    \midrule
    BSRGAN~\cite{bsrgan}        & \num{0.145707} & \num{0.131405} \\
    SwinIR~\cite{swinir}        & \num{0.035362} & \num{0.073155} \\
    HAT-S~\cite{HAT}            & \underline{\num{0.028607}} & \num{0.073064} \\
    RGT~\cite{rgt}              & \num{0.038265} & \num{0.074580} \\
    RGT-S~\cite{rgt}            & \num{0.040890} & \num{0.075927} \\
    SMFANet~\cite{smfanet}      & \num{0.037725} & \num{0.081912} \\
    SMFANet-plus~\cite{smfanet} & \num{0.035549} & \num{0.076578} \\
    ART-S~\cite{art}            & \num{0.031268} & \num{0.075017} \\
    PFT-light~\cite{pft-sr}        & \num{0.036361} & \num{0.075928} \\
    ATD-light~\cite{atd}        & \num{0.035944} & \underline{\num{0.072996}} \\
    HiT-SR~\cite{hitsr}         & \num{0.034698} & \num{0.077169} \\
    CoRPLE~\cite{corple}        & \num{0.037310} & \num{0.076598} \\
    \midrule
    \textbf{RPT-SR (Ours)}  & \textbf{\num{0.028481}} & \textbf{\num{0.072760}} \\
    \bottomrule
  \end{tabularx}
  \caption{Comparison of LPIPS scores on M3FD and RASMD datasets ($\times2$).}
  \label{tab:ablation-rasmdx2}
\end{table}

\subsection{Experimental Settings}
\label{sec:exp_settings}
For a fair comparison, we follow the training and evaluation protocols of recent state-of-the-art methods \cite{corple, difiisr}. We evaluate our model on both Long-Wave Infrared (LWIR) and Short-Wave Infrared (SWIR) super-resolution tasks. For LWIR SR, we use the M3FD \cite{m3fd} and TNO \cite{tno} datasets. M3FD is randomly split into 182 images for training and 38 for testing, while 37 images from TNO are used to assess cross-dataset generalization. For SWIR SR, we utilize the RASMD dataset \cite{rasmd}, partitioned into 3696 training and 498 test images. Following standard protocols, we generate low-resolution (LR) inputs by applying bicubic downsampling to high-resolution (HR) $512 \times 512$ image patches, creating $128 \times 128$ inputs for the $\times4$ task and $256 \times 256$ for the $\times2$ task. All $\times4$ experiments were conducted on an NVIDIA RTX 4090 GPU, while $\times2$ experiments were performed on an NVIDIA A6000 GPU.

\subsection{Comparison with State-of-the-Art Methods}
We conduct a comprehensive evaluation to benchmark our RPT-SR against recent state-of-the-art (SOTA) methods. Our evaluation encompasses both quantitative metrics and qualitative visualizations to thoroughly validate the superiority of our approach.

For quantitative analysis, we first compare performance on the M3FD dataset for the $\times4$ task. As presented in Table \ref{tab:m3fd}, RPT-SR sets a new state-of-the-art, achieving the best scores on LPIPS (0.104$\downarrow$) and MANIQA (0.262$\uparrow$). This indicates that our super-resolved images are not only perceptually closer to the ground truth but are also assessed as higher quality. In terms of model complexity, RPT-SR maintains a practical footprint (237.78G FLOPs), highlighting an effective balance between computational efficiency and performance.


To further validate the robustness and generalization capabilities of RPT-SR, we extend our quantitative analysis to additional datasets and magnification scales. For the $\times4$ task, as detailed in Table \ref{tab:ablation-rasmd}, our model exhibits highly competitive performance on both the RASMD and TNO datasets, achieving LPIPS scores on par with top-performing methods. When evaluated on the $\times2$ super-resolution task, RPT-SR again demonstrates exceptional performance. As shown in Table \ref{tab:ablation-rasmdx2}, our model sets a new state-of-the-art on the RASMD dataset and achieves a highly competitive second-best score on the M3FD dataset, closely following the leading method. This strong performance across different scales and diverse datasets underscores the versatility and effectiveness of our proposed architecture.

Complementing our quantitative findings, Figure \ref{fig:vis} provides a visual comparison on the M3FD dataset. The qualitative results clearly demonstrate our model's superior capability in reconstructing intricate details and textures. For instance, RPT-SR excels at preserving the structural integrity of human silhouettes where competing methods often introduce blurring artifacts (row 1). It adeptly restores sharp geometric patterns on building facades without the ringing or over-sharpening artifacts prevalent in other approaches (rows 2 and 3). Furthermore, in challenging low-contrast scenes, our model effectively mitigates noise amplification while reconstructing the object's original shape with higher fidelity (row 4).

Collectively, these comprehensive quantitative and qualitative evaluations robustly validate the efficacy of RPT-SR. Our model not only achieves state-of-the-art perceptual scores but also produces visually superior results across diverse and challenging conditions.

\begin{table*}[ht!]
    \centering
    \footnotesize
    \resizebox{\textwidth}{!}{%
        \begin{tabular}{l|ccc|ccc|ccc|c|c}
            \hline
            \multirow{2}{*}{Method} & \multicolumn{3}{c|}{M3FD x4} & \multicolumn{3}{c|}{TNO x4} & \multicolumn{3}{c|}{RASMD x4 (SWIR)} & \multirow{2}{*}{FLOPS} & \multirow{2}{*}{PARAMS} \\
             & LPIPS $\downarrow$ & MUSIQ $\uparrow$ & MANIQA $\uparrow$ & LPIPS $\downarrow$ & MUSIQ $\uparrow$ & MANIQA $\uparrow$ & LPIPS $\downarrow$ & MUSIQ $\uparrow$ & MANIQA $\uparrow$ & & \\
            \hline
            Baseline & 0.105057 & 41.2597 & 0.26091 & 0.25048 & 36.4154 & 0.24596 & 0.154442 & 44.39395 & 0.223607 & 231.366G & 25.454M \\
            Static & 0.105324 & 41.6872 & 0.26034 & 0.25257 & 38.2838 & 0.24903 & 0.154457 & 44.72269 & 0.225821 & 229.100G & 25.377M \\
            \textbf{RPT (Ours)} & \textcolor{red}{\textbf{0.103793}} & \textcolor{red}{\textbf{41.8049}} & \textcolor{red}{\textbf{0.26213}} & \textcolor{red}{\textbf{0.25014}} & \textcolor{red}{\textbf{38.5862}} & \textcolor{red}{\textbf{0.25125}} & \textcolor{red}{\textbf{0.153543}} & \textcolor{red}{\textbf{44.91027}} & \textcolor{red}{\textbf{0.226634}} & 237.783G & 25.830M \\
            \hline
        \end{tabular}%
        
    }
\caption{Ablation study of the proposed Regional Prior Transformer (RPT). We evaluate the impact of using only local tokens (\textbf{Baseline}), only static prior tokens (\textbf{Static}), and their fusion (\textbf{RPT (Ours)}). Best results are highlighted in bold \textcolor{red}{\textbf{red}}.}
\label{tab:ablation}
\end{table*}

\subsection{Ablation Study} 
\label{sec:ablation} 
\begin{figure}[ht!] 
    \includegraphics[width=\columnwidth]{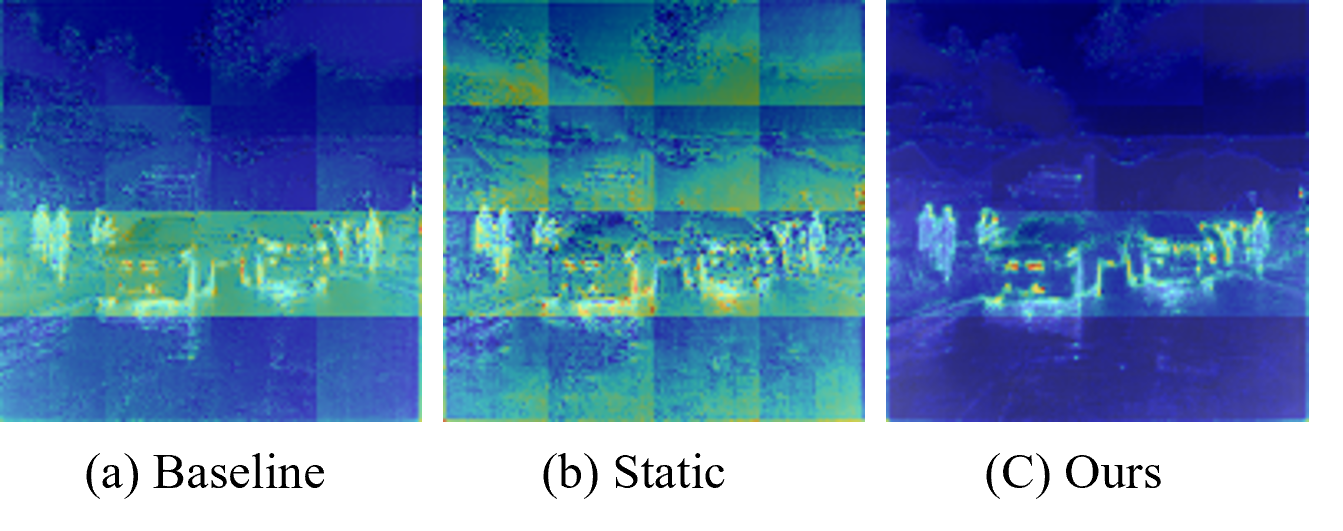} 
    \caption{Attention maps at the last RPA layer from the M3FD test set. From left to right we show the local-only baseline, the static-prior variant, and our full model.} 
    \label{fig:vis_att} 
\end{figure} 
To validate the effectiveness of the proposed components, we conduct an ablation study on the M3FD, TNO, and RASMD datasets. We compare our full Regional Prior Transformer (RPT) with two variants: (1) a \textit{baseline} model that uses only local tokens for attention (no regional priors), and (2) a \textit{static} model that uses only the learnable regional prior tokens (no dynamic local tokens). Table~\ref{tab:ablation} shows that both token types are beneficial and their combination yields the best performance across all datasets. For example, on M3FD our full RPT achieves a lower LPIPS and higher MUSIQ and MANIQA scores than the baseline, indicating that learned regional priors provide complementary contextual information beyond what can be obtained from local content alone. The static-only model, which relies purely on positional priors, improves some metrics (\eg MUSIQ on TNO) but consistently lags behind the full RPT, confirming that priors without frame-specific content are insufficient to reconstruct fine details of the current image. Figure~\ref{fig:vis_att} further illustrates these behaviors by visualizing the attention map of the three variants. The baseline produces a coarse horizontal band of attention around the horizon, while the static-only model exhibits diffuse checkerboard patterns. In contrast, RPT concentrates attention on semantically meaningful regions such as vehicles, and pedestrians, while suppressing sky and background responses. Together with the quantitative gains, these visualizations support our hypothesis that the best performance is obtained when frame-specific local information is modulated by persistent, scene-constant regional priors. Importantly, this improvement is achieved with only a marginal increase in FLOPs compared to the baseline, demonstrating the efficiency of the proposed design.

\section{Conclusion}
We addressed the inefficiency of general-purpose super-resolution models in fixed-viewpoint infrared imaging by explicitly exploiting stable spatial priors. To this end, we proposed RPT-SR, a Regional Prior attention Transformer with a dual-token mechanism that fuses frame-specific local tokens with learnable regional prior tokens encoding the scene layout. Experiments on both Long-Wave and Short-Wave infrared datasets show that RPT-SR consistently achieves state-of-the-art perceptual performance, indicating that the model effectively learns scene-constant structural regularities beyond the specifics of the imaging physics. This gain comes at the cost of a modest increase in parameters due to the additional prior tokens, which we view as a deliberate trade-off between model size and perceptual quality. Future work includes compressing these priors and extending the regional prior concept to other restoration tasks such as video super-resolution.

\section{Acknowledgement}
This work was supported by Institute of Information \& communications Technology Planning \& Evaluation (IITP) grant funded by the Korea government(MSIT) (RS-2025-02218237, Development of Digital Innovative Technologies for Enhancing the Safety of Complex Autonomous Mobility)

{
    \small
    \bibliographystyle{ieeenat_fullname}
    \bibliography{main}
}

\end{document}